%% file: main.tex
\documentclass{article}

\usepackage{PRIMEarxiv}

\usepackage[utf8]{inputenc}
\usepackage[T1]{fontenc}
\usepackage{hyperref}
\usepackage{url}
\usepackage{graphicx}
\usepackage{booktabs}
\usepackage{multirow}
\usepackage{amsmath,amssymb}
\usepackage{amsfonts}
\usepackage{nicefrac}
\usepackage{microtype}
\usepackage{xcolor}
\usepackage{algorithm}
\usepackage{algorithmic}
\usepackage{subcaption}
\usepackage{wrapfig}

\graphicspath{{figs/}}

\title{Neural Architecture Distributions: A New Paradigm for Stochastic Segmentation}

\author{
  Conghui Li\textsuperscript{1},
  Junhao Huang\textsuperscript{2},
  Chern Hong Lim\textsuperscript{1},
  Bing Xue\textsuperscript{2},
  Mengjie Zhang\textsuperscript{2}
  \\
  \textsuperscript{1}School of Information Technology, Monash University Malaysia, Selangor, Malaysia
  \\
  \textsuperscript{2}Centre for Data Science and Artificial Intelligence, School of Engineering and Computer Science,
  \\
  Victoria University of Wellington, Wellington, New Zealand
  \\
  \texttt{Conghui.Li@monash.edu, Lim.ChernHong@monash.edu}
  \\
  \texttt{junhao.huang@vuw.ac.nz, Bing.Xue@vuw.ac.nz, mengjie.zhang@ecs.vuw.ac.nz}
}

\begin{document}
\maketitle
\begin{abstract}
    Stochastic segmentation seeks to represent multiple plausible masks for a single image, which is essential in safety- and quality-critical applications such as medical imaging or building defect inspection. Most existing methods introduce stochasticity by injecting continuous latent variables or by iterative denoising trajectories, whose stochastic sources are difficult to search or audit directly. We propose architecture distributions as a new stochastic source for segmentation: instead of sampling a latent variable or noise, we sample a discrete architecture from a learned distribution over operator choices at multiple searchable positions in a segmentation backbone. Each sampled architecture yields one mask through the selected active path, so inference depends on the executed subnet rather than the complete candidate bank. This approach also supports architectural provenance, since each output corresponds to a specific architecture configuration. To reduce collapse toward averaged masks, we train with set-level supervision by matching a set of architecture-sampled predictions to the annotation set using an IoU-based energy-distance surrogate. We further construct the candidate bank with evolutionary search, making the support of the stochastic source optimizable before distribution learning. The proposed method achieves state-of-the-art distribution matching and hypothesis coverage on LIDC-IDRI, and remains effective on two extension tasks. To the best of our knowledge, this is the first work to formulate stochastic segmentation as learning an architecture distribution and realizing output diversity through architecture sampling.
\end{abstract}


\section{Introduction}
Semantic segmentation is a fundamental component in many safety- and quality-critical vision systems, including medical image analysis, building and infrastructure inspection, and transportation-related perception~\cite{Kohl2018}. In these applications, the target mask is often ambiguous: multiple segmentations can be plausible for the same input from different experts' annotations due to limited image evidence, weak boundaries, or inherent subjectivity among annotators~\cite{Zbinden2023,Monteiro2020}. Therefore, beyond producing a single point estimate, it is desirable to model a conditional distribution over masks given an image, so that the model can generate multiple plausible hypotheses and support uncertainty-aware decision making.

Most modern segmentation pipelines are built upon deterministic convolutional backbones and typically output one mask per input. Architectures such as U-Net and its variants have become standard due to their competitive empirical performance and efficient inference~\cite{Ronneberger2025, Zhou2018, Hu2022}. However, when ambiguity exists, pixel-wise supervision with a single annotation encourages the model to predict an ``average'' solution, which may not correspond to any valid annotation mode. This limitation motivates stochastic segmentation, where the goal is to approximate the distribution of plausible masks rather than a single output.

Our motivation is that many segmentation ambiguities are not arbitrary pixel-wise noise, but discrete structural alternatives. For example, annotators may disagree on whether a weak boundary should be included, whether a small uncertain component is connected to the main region, or how far an object should extend into low-contrast areas. Such alternatives can be viewed as different structural priors over boundary sharpness, receptive field, local smoothness, and shape connectivity. Candidate operators in an architecture bank naturally encode these biases through their kernel size, dilation, depth, residual/MBConv/ConvNeXt-style transformations, and identity paths. Sampling an architecture therefore samples an executable decision function with a particular structural bias, rather than only perturbing the feature representation of a fixed decoder. This motivates using an input-conditioned architecture distribution to represent annotator-dependent hypotheses.

Existing stochastic segmentation approaches commonly introduce randomness through latent variables or iterative generative processes. A representative line of work couples a segmentation backbone with conditional latent-variable models, such as the Probabilistic U-Net~\cite{Kohl2018}, to sample diverse outputs by drawing a latent code. Other approaches model a structured distribution directly in the logit space, such as Stochastic Segmentation Networks (SSN)~\cite{Monteiro2020}. More recently, diffusion and score-based generative models have been explored for image generation and have inspired stochastic prediction mechanisms that rely on iterative denoising or numerical solvers \cite{Rahman2023,Zbinden2023}. To increase the expressiveness of latent distributions, several works also augment variational posteriors using normalizing flows, which transform simple distributions through sequences of invertible mappings \cite{Conghui2025,Papamakarios2021,Valiuddin2021}.

Despite their progress, latent-variable and diffusion-style stochastic segmentation methods have two practical drawbacks for deployment in safety-critical settings. First, improving the expressiveness of the stochastic source often increases inference cost: diffusion-type models typically require multiple sampling steps, and flow-augmented variational models add extra transformations to obtain a richer posterior \cite{Ho2020,Papamakarios2021}. Second, sample traceability is limited. Although the posterior, decoder, or score network is optimized, the underlying randomness is usually represented by a continuous noise/code or denoising trajectory, so individual masks are not naturally associated with a discrete, human-interpretable generation pathway. In safety-critical settings, it is often valuable to record how a specific hypothesis is produced, enabling auditing and diagnosis when predictions are used to support decisions.

In this work, we propose a different perspective for dealing with stochastic segmentation: instead of injecting stochasticity into features or latent variables, we define the model itself as a distribution over architectures. Concretely, we insert a set of searchable positions into a backbone and define a categorical architecture distribution over candidate operators at each position. This turns the stochastic source into a discrete, searchable support: the candidate bank can be shaped by evolutionary search before the input-conditioned distribution is learned with set-level supervision. At inference time, one segmentation sample is obtained by sampling an architecture and executing its active path. This design has two advantages. First, inference is determined by the selected active path rather than by the complete stored candidate bank, because only one operator per searchable position is executed for each sample. Second, every predicted mask is associated with an explicit discrete architecture code, providing a transparent and auditable generation trace.

To learn a distribution of plausible masks, we adopt set-level supervision that matches a set of predicted samples to the set of annotations, and we optimize a distributional objective based on the generalized energy distance as popularized for stochastic segmentation evaluation \cite{Kohl2018}. We further construct a performance-oriented candidate operator bank using an aging-evolution style search procedure \cite{Real2019}, and train the final architecture distribution model with weight inheritance and stochastic architecture sampling. In summary, our contributions are:

(1). We introduce architecture distributions as a new stochastic source for segmentation, where randomness is realized by sampling discrete architectures rather than injecting latent noise.

(2). We show that the support of this stochastic source can be optimized by evolutionary construction of the candidate bank, providing a searchable alternative to fixed latent/noise sources.

(3). We develop a set-level objective based on an energy-distance surrogate to prevent distribution degradation (collapse or averaging) and to better match multi-annotator labels.

(4). We demonstrate that architecture sampling improves distribution matching while maintaining competitive inference time and providing architectural provenance.

\section{Related Work}
\subsection{Stochastic Segmentation}
Stochastic segmentation aims to model multiple plausible masks for the same input, rather than a single deterministic output. This setting is common when boundaries are visually ambiguous, when multiple annotators disagree, or when the target concept is inherently subjective. Early uncertainty baselines often rely on approximate Bayesian inference like Monte-Carlo dropout or ensembles to quantify predictive uncertainty~\cite{Gal2016, Lakshminarayanan2017}; however, these approaches are not designed to match the empirical distribution induced by multi-annotator labels and may still favor an “average” solution.

A major line of work formulates stochastic segmentation as conditional generative modeling with latent variables. Probabilistic U-Net combines a U-Net backbone with a Conditional Variational Autoencoder (CVAE) to sample diverse segmentations conditioned on the image, and evaluates distribution matching with set-level metrics such as generalized energy distance~\cite{Kohl2018}. PHiSeg further introduces hierarchical latent variables to capture ambiguities across multiple spatial scales, addressing the limitation that a single global latent code may be ignored or may not express localized variations~\cite{Baumgartner2019, Simon2019}. Follow-up works use flow-based methods to better align the prior and posterior latent distributions, giving the latent variables richer semantic structure~\cite{Selvan2020, Valiuddin2021, Bhat2022, De2025}.

Another direction represents stochasticity directly in the segmentation field or logit space. SSN models a distribution over segmentation logits with spatial correlation structure and draws samples that yield coherent masks without relying on an explicit latent-image decoder~\cite{Monteiro2020}. More recently, diffusion models have been explored for generating segmentation maps via iterative denoising. Many of them adapts denoising diffusion to this task and demonstrates competitive results on general benchmarks~\cite{Rahman2023, Zbinden2023, Baru2026}. However, diffusion sampling typically requires many iterative steps, which can be computationally expensive at inference.

\subsection{Architecture Distributions}
Learning distributions over discrete architectural choices has been extensively studied in neural architecture search (NAS). Differentiable NAS (DARTS)~\cite{Hanxiao2019} relaxes discrete operator selection into continuous mixtures, enabling gradient-based optimization. Stochastic NAS (SNAS)~\cite{Sirui2019} models operator choices as categorical random variables and uses reparameterized sampling for end-to-end learning of an architecture distribution. Supernet-based methods (ENAS)~\cite{Pham2018} train a single over-parameterized network and evaluate many sub-networks by sampling architectures that inherit shared weights. Once-for-All (OFA)~\cite{Han2020} extends this idea by training a single network that supports a very large family of subnetworks, enabling fast specialization by subnetwork selection.

These NAS methods typically learn architecture distributions to search for a single best model under accuracy and efficiency constraints. In contrast, stochastic segmentation focuses on learning a conditional output distribution with multiple plausible masks. This motivates us using architecture sampling not only as a search tool, but as a mechanism to parameterize and sample from the model’s conditional prediction distribution.

\begin{figure}
	\centering
\includegraphics[width=12 cm]{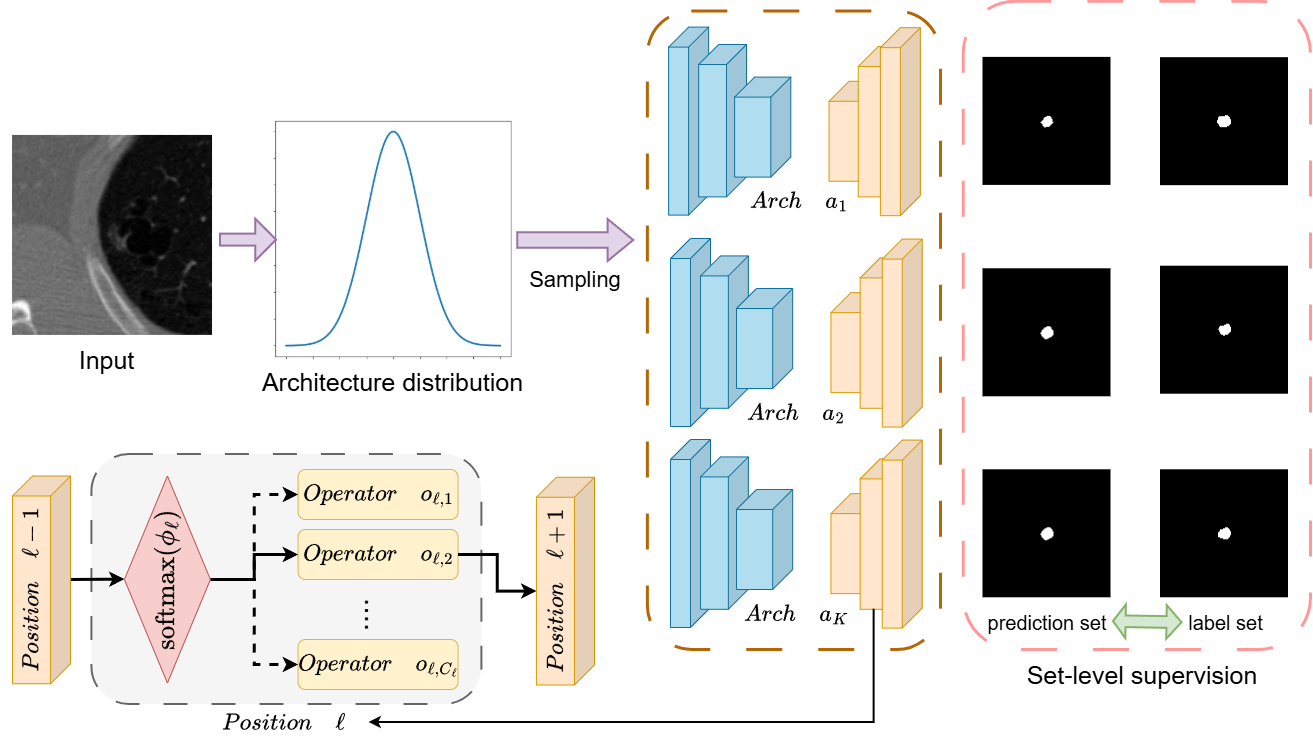}
\caption{Overview of Architecture Distribution Sampling framework. Given an input, we predict an input-conditioned distribution over discrete architectures, from which we sample (K) architectures. Each sampled architecture instantiates a single-path subnet to produce a set of segmentation hypotheses, which is trained via set-level supervision by matching the prediction set to the multi-annotator label set.}
\label{methodology}
\end{figure}

\section{Methodology}

\subsection{Preliminaries}

We study stochastic segmentation where multiple segmentation masks can be valid for the same input image. Let $x \in \mathbb{R}^{C_{\mathrm{in}}\times H \times W}$ denote an input image. For each image, we observe a \emph{set} of $M$ binary annotations
\begin{equation}
Y(x)=\{y^{(m)}\}_{m=1}^{M}, \qquad y^{(m)} \in \{0,1\}^{H\times W},
\end{equation}
provided by different annotators or repeated annotations. This setting corresponds to an unknown conditional distribution $p(y\mid x)$ rather than a single deterministic target.

Our goal is to learn a model that can produce multiple plausible segmentations for a given input. At inference time, the model should generate a set of $K$ predictions (where $K \geq M$).

\begin{equation}
\hat{\mathcal{Y}}(x)=\{\hat{y}^{(1)},\hat{y}^{(2)},\dots,\hat{y}^{(K)}\}
\end{equation}

such that the distribution of $\hat{\mathcal{Y}}(x)$ matches the annotation distribution represented by $\mathcal{Y}(x)$.

Unlike latent-variable approaches (CVAE or diffusion based methods) that inject random noise into the model inputs or hidden states, we represent stochasticity as a distribution over network architectures. Concretely, we define a discrete architecture variable $a$ and a learnable, input-conditioned architecture distribution $q_{\psi}(a\mid x)$. Each sample $a \sim q_{\psi}(a\mid x)$ specifies a concrete architecture instance within a weight-inheritance network, and produces a prediction through

\begin{equation}
\hat{p}(x;a)=\sigma(f_{\theta}(x;a)), \qquad \hat{y}(x;a)=\mathbb{I}[\hat{p}(x;a) > 0.5]
\end{equation}
where $f_{\theta}(\cdot;a)$ denotes the segmentation network instantiated by architecture $a$, $\theta$ are shared weights, and $\sigma(\cdot)$ is the sigmoid function. For multi-class segmentation, we replace the sigmoid head with a softmax head. Therefore, a set of predictions is obtained by sampling architectures multiple times:

\begin{equation}
a^{(k)} \sim q_{\psi}(a\mid x), \quad \hat{y}^{(k)} = \hat{y}(x; a^{(k)}), \quad k=1,\dots,K.
\end{equation}

This formulation makes the model output a segmentation distribution through architecture sampling, while keeping the underlying weights inherited and the forward pass standard for each sampled architecture.

\subsection{Overall architecture}

We propose an architecture-distribution segmentation network, which we call NADS (Neural Architecture Distribution Sampling), that models ambiguity by sampling discrete architectural instances at inference time. The framework is shown in Fig.~\ref{methodology}. Unlike CVAE- or diffusion-based stochastic segmentation, which introduces randomness through an explicit latent variable (sampling $z \sim p(z)$ and generating $\hat{y}$ via $p_{\theta}(y\mid x,z)$ or an iterative noise-to-mask process), our method places stochasticity directly on the model structure by sampling an architecture $a \sim q_{\psi}(a\mid x)$ and producing predictions through a deterministic forward pass $ \hat{y}(x;a)=\mathbb{I}[\sigma(f_{\theta}(x;a))>0.5] $, so diversity is generated by architecture sampling rather than latent-noise injection.

\subsubsection{Architecture variable}

We define a segmentation network that contains $L$ searchable positions, each corresponding to one SearchableBlock inserted into the backbone (e.g., several blocks in the decoder). At each position $\ell \in {1,\dots,L}$, the network provides a finite set of candidate operators, $\mathcal{O}_{\ell}={o_{\ell,1},o_{\ell,2},\dots,o_{\ell,C_{\ell}}}$, where each $o_{\ell,c}$ is a shape-preserving mapping that takes a feature map and returns a feature map with the same tensor shape. This ensures that all operator choices are compatible with the same backbone topology. 

The architecture variable is a collection of discrete choices $a=(a_1,a_2,\ldots,a_L)$, where each $a_\ell \in \{1,\ldots,C_\ell\}$ selects one candidate operator at position $\ell$. Intuitively, sampling $a$ instantiates a concrete architecture by specifying which candidate operator is active at each searchable position. Importantly, the global network layout remains unchanged; only the internal operator choices at these $L$ positions vary across samples.

\subsubsection{Architecture distribution}

We learn a conditional probability distribution over architectures $q_{\psi}(a \mid x)$, parameterized by an architecture controller $g_{\psi}$ that produces input-dependent logits from image features. Specifically, given an input image $x$, we extract a feature representation $h(x)$ from the backbone encoder, and the controller outputs $\{s_{\ell}(x)\}_{\ell=1}^{L}$, where $s_{\ell}(x)\in\mathbb{R}^{C_{\ell}}$ controls the selection probability among $\mathcal{O}_{\ell}$. We assume the conditional distribution factorizes across positions:

\begin{equation}
q_{\psi}(a\mid x)=\prod_{\ell=1}^{L} q_{\psi}(a_\ell\mid x),
\qquad
q_{\psi}(a_\ell=c\mid x)=\big[\mathrm{softmax}(s_\ell(x))\big]_c
\end{equation}
Here, $s_\ell(x)=g_{\psi}(h(x))_\ell$. This factorization avoids enumerating an exponential number of full architectures, while still representing a large space for efficient sampling and learning. Diversity arises from composing different local operator choices across positions conditioned on the input.

Although factorized, the induced conditional distribution over full architectures remains expressive because the number of possible architectures is $|\mathcal{A}| = \prod_{\ell=1}^{L} C_{\ell}$.

\subsubsection{Weight inheritance}

We denote the weight-inheritance network by $f_{\theta}$. Given a sampled architecture $a$, the network produces logits $z(x;a) = f_{\theta}(x;a),$ and probabilities $p(x;a) = \sigma(z(x;a))$.
Crucially, the randomness only enters through $a$. Conditioned on $a$, the forward pass is deterministic. All architectures inherit the same parameter set $\theta$, but activate different operator subsets (and therefore different parameter subsets) inside the shared supernetwork.

\subsubsection{Training \& Inference}

At inference time, stochastic segmentations are generated by sampling architectures:
\begin{equation}
a^{(k)} \sim q_{\psi}(a\mid x), \qquad \hat{y}^{(k)} = \mathbb{I}[\sigma(f_{\theta}(x; a^{(k)})) > 0.5], \quad k=1,\dots,K.
\end{equation}

The prediction set $\{\hat{y}^{(k)}\}_{k=1}^{K}$ approximates the model's conditional output distribution, enabling evaluation with distributional metrics and supporting downstream uncertainty estimation.

Direct optimization of discrete architectural choices is not differentiable. During training, we use a differentiable relaxation to update $\psi$, while still training the inherited weights $\theta$ with sampled architectures. Specifically, at each searchable position we draw a relaxed one-hot vector $\tilde{a}_\ell$ using a straight-through Gumbel-Softmax distribution~\cite{Eric2017} with temperature $\tau$:

\begin{equation}
\tilde{a}_\ell = \mathrm{GumbelSoftmax}(s_\ell(x),\tau).
\end{equation}

This relaxation enables gradient-based learning of $\psi$. In later stages, we progressively reduce $\tau$ to encourage sharper distributions and more discrete behavior.

\subsection{Objective function}

A key challenge in our setting is that pixel-wise supervision against a single annotation can drive the model toward an ``average'' mask that does not correspond to any plausible annotator output. To learn a distribution of valid masks, we employ set-level supervision: a set of sampled predictions is trained to match the set of available annotations for each image.

\subsubsection{GED surrogate loss}
For the $k$-th sampled architecture $a^{(k)} \sim q_{\psi}(a\mid x)$, we denote the predicted
probability map as

\begin{equation}
p^{(k)}(x) \triangleq \hat p(x; a^{(k)}) = \sigma\!\big(f_\theta(x; a^{(k)})\big).
\end{equation}
We adopt sample-based energy distance with a soft-IoU ground metric. We use the soft-IoU distance

\begin{equation}
d_{\mathrm{IoU}}(u,v)
\triangleq 1 - \frac{\langle u,v\rangle}
{\|u\|_1+\|v\|_1-\langle u,v\rangle+\epsilon},
\end{equation}
where $u\in[0,1]^{H\times W}$ and $v\in[0,1]^{H\times W}$. Then we compute the following averages:

\begin{align}
\bar d_{PY}(x) &= \frac{1}{KM}\sum_{k=1}^{K}\sum_{m=1}^{M} d_{\mathrm{IoU}}\!\left(p^{(k)}(x),y^{(m)}(x)\right),\\
\bar d_{PP}(x) &= \frac{1}{K(K-1)}\sum_{\substack{k,k'=1\\k\neq k'}}^{K} d_{\mathrm{IoU}}\!\left(p^{(k)}(x),p^{(k')}(x)\right)
\end{align}

Our set-level loss is an energy-distance surrogate. Relative to the classic GED, we drop the annotation-only term $\bar d_{YY}$ (constant w.r.t.\ model parameters), reweight the diversity term by $\beta$, and clip the result at zero to preserve non-negativity:

\begin{equation}
\mathcal{L}_{\mathrm{set}}(x)
=
\max\!\Big(0,\; 2\,\bar d_{PY}(x) - \beta\,\bar d_{PP}(x) \Big),
\end{equation}
where $\beta$ controls the strength of the diversity term among predictions. The cross term $\bar d_{PY}$ encourages sampled predictions to be close to the annotation set, while the negative self term $-\beta\,\bar d_{PP}$ discourages mode collapse by penalizing overly similar predictions. 

\subsubsection{Full objective}
In addition, we add an auxiliary stabilizer acting on the mean prediction and the mean annotation to ease early optimization. Let $z^{(k)}(x)=f_\theta(x;a^{(k)})$ denote the logit map produced by the $k$-th sampled architecture.

\begin{equation}
\bar z(x)=\frac{1}{K}\sum_{k=1}^{K}z^{(k)}(x),
\qquad
\bar y(x)=\frac{1}{M}\sum_{m=1}^{M}y^{(m)}(x).
\end{equation}
We define $\mathcal{L}_{\mathrm{aux}}(x)$ as a standard BCE--Dice loss applied to $\bar z(x)$ and $\bar y(x)$. This term is intended only as a stabilizer: with $\mathcal{L}_{\mathrm{set}}$ kept as the dominant objective through a small $\lambda_{\mathrm{aux}}$, it does not drive the model toward a single averaged mask.
Finally, we regularize the architecture distribution using negative entropy to avoid premature collapse:

\begin{equation}
\mathcal{R}(q_{\psi})
= -\,\mathbb{E}_{x}\sum_{\ell=1}^{L} H\!\left(q_{\psi}(a_\ell\mid x)\right)
= \mathbb{E}_{x}\sum_{\ell=1}^{L}\sum_{c=1}^{C_\ell} q_{\psi}(a_\ell=c\mid x)\,
\log\!\big(q_{\psi}(a_\ell=c\mid x)+\varepsilon\big).
\end{equation}
where $q_{\psi}(a_\ell\mid x)$ is the input-conditioned categorical distribution at position $\ell$. The full training objective over a batch with weights $\lambda_{\mathrm{aux}}$ and $\lambda_{\mathrm{arch}}$ is

\begin{equation}
\mathcal{L}
=
\mathbb{E}_{x}\Big[\mathcal{L}_{\mathrm{set}}(x) + \lambda_{\mathrm{aux}}\mathcal{L}_{\mathrm{aux}}(x)\Big]
+\lambda_{\mathrm{arch}}\mathcal{R}(q_{\psi})
\end{equation}

\subsection{Variable architecture search}

In our framework, stochasticity comes from sampling among discrete candidate blocks at each searchable position. We construct a position-specific candidate bank using an efficient evolutionary procedure, and then learn the architecture distribution over the retained candidates.

\subsubsection{Search space}

We define $L$ searchable positions in the baseline network. At each position $\ell$, a candidate block is specified by a compact genotype that encodes the internal micro-architecture of a shape-preserving module. All candidates are constrained to satisfy $o_{\ell,c}: \mathbb{R}^{C\times H\times W}\rightarrow \mathbb{R}^{C\times H\times W},$ so they can be swapped without changing tensor shapes or the global backbone topology.

\subsubsection{Aging evolution}
We construct the candidate operator bank for each searchable position using aging evolution. Specifically, we maintain a fixed-size population of candidate genotypes. At each iteration, we randomly sample a small subset of the population, select the best-performing individual as the parent, generate a child by applying a random mutation, and evaluate the child under a fixed, budgeted training protocol using the same stochastic-segmentation objective. The child is inserted into the population and the oldest individual is removed to preserve constant population size, which encourages exploration and reduces premature convergence. After the search, we retain the top-$N$ candidates at each position to form the final candidate bank, and learn the architecture distribution over this bank during end-to-end training.

\subsubsection{Fitness evaluation}

We lock the supernet to the genotype, which restricts the candidate operator set at each position, and estimate HM-IoU/GED using a small number of stochastic forward passes by sampling architectures from the induced distribution. We combine them into a scalar objective

\begin{equation}
\mathrm{fit}(g) = (1-\mathrm{HM\text{-}IoU}) + \lambda_{\mathrm{ged}}\mathrm{GED} + \lambda_{\mathrm{cost}}\mathrm{Cost}(g),
\end{equation}
where $Cost(g)$ is a complexity proxy accumulated from the selected genes. Including this proxy biases the search toward lower-cost candidate structures, reducing the computational cost of each sampled active path at inference.

\section{Experiment}

\subsection{Main Experiments}
We design three different experiments to validate our method from complementary perspectives: the main experiment, stochastic segmentation on LIDC-IDRI with multi-annotator supervision, where modeling a conditional output distribution is essential; multi-class deterministic segmentation on Cityscapes, to verify that our architecture-distribution training does not sacrifice standard semantic segmentation accuracy~\cite{Cordts2016}; and single-class curvilinear deterministic segmentation on Crack500, to test robustness on thin-structure targets with severe class imbalance~\cite{Zhang2016}. Unless otherwise stated, we follow the dataset protocols from ~\cite{Conghui2025} and~\cite{Chen2024} for fair comparison.
For deterministic datasets, the architecture distribution is not used for distribution matching. Instead, the controller is trained to minimize the expected standard segmentation loss, and typically converges to a near-deterministic policy. At inference, we use Maximum a Posteriori (MAP) inference for efficient prediction. In addition, when only a single annotation is available per image ($M=1$), distribution matching at the set level is not identifiable. Therefore, we train using the standard Dice loss.

\subsubsection{Evaluation metrics.}
For LIDC, we evaluate distribution matching using GED and HM-IoU, both computed between a set of $K$ predicted samples and the annotation set. We report GED$_K$ and HM-IoU$_K$ under multiple $K$ values to reflect the quality of the learned conditional distribution as the number of samples increases. For Cityscapes and Crack500, we report standard single-output segmentation metrics: IoU on Cityscapes, and Precision, Recall, and F1-score on Crack500.

\begin{table*}[ht]
\caption{Quantitative results on LIDC-IDRI, with methods listed in chronological order. All baseline results are taken from~\cite{Conghui2025}. Our results are over 3 seeds. Note that lower values of GED indicate better performance, while higher values of HM-IoU are preferred.}
\centering
\resizebox{\textwidth}{!}{%
\begin{tabular}{lcccccc}
\toprule
\textbf{Method} & \textbf{GED$_{16}$} & \textbf{GED$_{32}$} & \textbf{GED$_{50}$} & \textbf{GED$_{100}$} & \textbf{HM-IoU$_{16}$} & \textbf{HM-IoU$_{32}$} \\
\midrule
Prob. Unet \cite{Kohl2018}       & 0.310$_{\pm0.01}$ & 0.303$_{\pm0.01}$ & --                   & 0.252$_{\pm0.004}$ & 0.552$_{\pm0.00}$ & 0.548$_{\pm0.00}$ \\
HProb. Unet \cite{Simon2019}     & 0.270$_{\pm0.01}$ & --                   & --                   & --                   & 0.530$_{\pm0.01}$ & -- \\
PhiSeg \cite{Baumgartner2019}             & 0.262$_{\pm0.00}$ & 0.247$_{\pm0.00}$ & --                   & 0.224$_{\pm0.004}$ & 0.586$_{\pm0.00}$ & 0.595$_{\pm0.00}$ \\
SSN \cite{Monteiro2020}                   & 0.259$_{\pm0.00}$ & 0.243$_{\pm0.01}$ & --                   & 0.225$_{\pm0.002}$ & 0.558$_{\pm0.00}$ & 0.555$_{\pm0.01}$ \\
cFlow \cite{Selvan2020} & --                   & 0.225$_{\pm0.01}$ & --                   & --                   & --                    & 0.584$_{\pm0.00}$ \\
CAR \cite{Kassapis2021}                   & --                   & --                   & --                   & 0.228$_{\pm0.009}$           & --                    & -- \\
JProb. Unet~\cite{Zhang20222}     & --                   & 0.206$_{\pm0.01}$     & --                   & --                   & --                    & 0.647$_{\pm0.01}$ \\
PixelSeg \cite{Zhang20221} & 0.243$_{\pm0.01}$     & --                   & --                   & --                   & 0.614$_{\pm0.01}$      & -- \\
MoSE \cite{Gao2022} & 0.218$_{\pm0.03}$     & 0.195$_{\pm0.002}$    & 0.195$_{\pm0.002}$   & 0.189$_{\pm0.002}$           & 0.624$_{\pm0.004}$ & -- \\
AB \cite{Chen2022} & 0.213$_{\pm0.01}$     & 0.196$_{\pm0.02}$     & 0.193$_{\pm0.002}$   & -- & 0.614$_{\pm0.01}$      & 0.619$_{\pm0.001}$ \\
CIMD \cite{Rahman2023} & 0.234$_{\pm0.005}$    & 0.218$_{\pm0.005}$    & 0.210$_{\pm0.005}$   & -- & 0.587$_{\pm0.01}$      & 0.592$_{\pm0.002}$ \\
CCDM \cite{Zbinden2023}           & 0.212$_{\pm0.002}$ & 0.194$_{\pm0.001}$ & 0.187$_{\pm0.002}$ & 0.183$_{\pm0.002}$ & 0.623$_{\pm0.002}$ & 0.631$_{\pm0.002}$ \\

GMM\_NF \cite{Conghui2025} & 0.196$_{\pm0.001}$ & 0.189$_{\pm0.001}$ & 0.185$_{\pm0.002}$ & 0.184$_{\pm0.000}$ & 0.641$_{\pm0.002}$ & 0.644$_{\pm0.001}$ \\

\midrule

Ours & \textbf{0.172$_{\pm0.001}$} & \textbf{0.169$_{\pm0.001}$} & \textbf{0.165$_{\pm0.003}$} & \textbf{0.164$_{\pm0.003}$} & \textbf{0.658$_{\pm0.001}$} & \textbf{0.671$_{\pm0.002}$} \\

\bottomrule
\end{tabular}%
}
\label{table_medical}
\end{table*}

\subsubsection{Stochastic Segmentation}
Table~\ref{table_medical} reports the quantitative results on LIDC, with methods listed in chronological order; all baseline values are taken from~\cite{Conghui2025}. Our approach achieves the best results on both distribution-matching and coverage-oriented metrics: it obtains the lowest GED under every sampling budget and the highest HM-IoU among all compared methods.

\begin{figure}
	\centering
\includegraphics[width=12 cm]{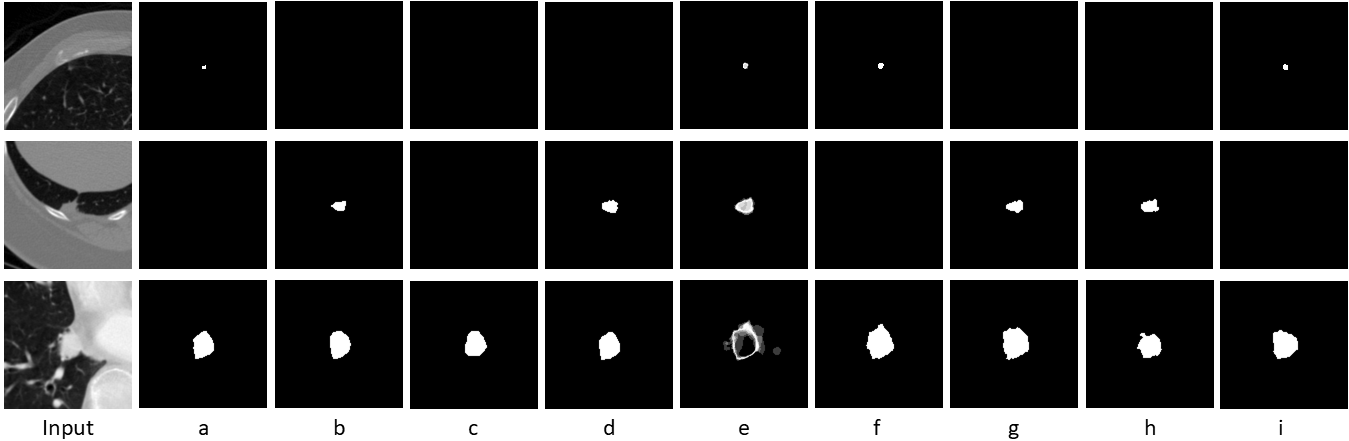}
\caption{Qualitative results on LIDC images with our method. (a)-(d) denote as ground truth label from four different experts. (e) represent the variance of 16 predictions. (f)-(i) represent the 4 of 16 prediction from proposed method.}
\label{lidc}
\end{figure}

Specifically, our method attains HM-IoU$_{16}=0.658$ and HM-IoU$_{32}=0.671$, exceeding the best previously reported baseline on each metric (HM-IoU$_{16}=0.641$ by GMM\_NF and HM-IoU$_{32}=0.647$ by JProb.\ U-Net) by $+1.7$ and $+2.4$ points, respectively. The improvement in GED is more pronounced: our method reduces GED from the strongest baseline values of 0.196, 0.189, 0.185, and 0.183 to 0.172, 0.169, 0.165, and 0.164 for $K=\{16,32,50,100\}$. Since GED penalizes both missing annotation modes and generating samples that are far from the annotator set, these gains suggest that architecture sampling does not merely increase diversity; it produces diverse hypotheses that remain well aligned with the empirical multi-rater distribution. Qualitative examples are shown in Fig.~\ref{lidc}. Our sampled predictions (f–i) produce multiple plausible masks that vary in size and shape in a way that mirrors these annotator differences, rather than collapsing to a single averaged contour.

\begin{table}[ht]
\centering
\begin{minipage}[t]{0.48\textwidth}
\centering
\caption{Quantitative results on Cityscapes. All baseline results are from~\cite{Conghui2025}. ``Param'' is the model parameter count in millions; for our method it is the activated MAP subnet, while the full supernet stores 387M.}
\begin{tabular}{lllcc}
\toprule
\textbf{Method} & \textbf{Backbone} & \textbf{Param} & \textbf{IoU} \\
\midrule
DeepLabv3 \cite{Chen2017}   & ResNet50   & 39M & 58.6 \\
DeepLabv3 \cite{Chen2017}   & ResNet101  & 58M & 59.2 \\
UPerNet \cite{Xiao2018}       & ResNet101  & 83M & 60.7 \\
HRNet \cite{Wang2020}           & w48v2      & 70M & 63.3 \\
UPerNet \cite{Liu2021}   & Swin-Tiny  & 58M & 65.5 \\
CCDM \cite{Zbinden2023}                  & -          & 30M & 60.3 \\
CCDM \cite{Zbinden2023}                  & Dino ViT-S & 50M & 65.8 \\
Prob. Unet \cite{Kohl2018}                 & -          & 33M & 63.2 \\
GMM\_NF \cite{Conghui2025}                  & - & 38M & 73.0 \\
\midrule
Ours                  & - & 26M & \textbf{74.2} \\
\bottomrule
\label{table_cityscape}
\end{tabular}
\end{minipage}
\hfill
\begin{minipage}[t]{0.48\textwidth}
\centering
\caption{Quantitative results on Crack500. All baseline results are from~\cite{Chen2024}. The proposed method obtains the highest precision and F1-score.}
\begin{tabular}{lccc}
\toprule
\textbf{Method} & \textbf{Precision} & \textbf{Recall} & \textbf{F1} \\
\midrule
UNet \cite{Ronneberger2025}              & 62.22 & 68.85 & 61.83 \\
VGG-UNet \cite{Mosinska2018}       & 58.18 & 60.26 & 51.79 \\
TopoNet \cite{Hu2019}        & 66.81 & 62.68 & 60.06 \\
DRU \cite{Wang2019}                & 61.94 & \textbf{71.43} & 62.82 \\
Crackformer \cite{Liu20212}& 69.13 & 66.24 & 64.75 \\
JTFN \cite{Cheng2021}              & 68.81 & 69.06 & 65.76 \\
JTFN + CIRL \cite{Chen2024}          & 70.32 & 69.93 & 67.62 \\
Prob. Unet \cite{Kohl2018}         & 56.80 & 69.39 & 60.80 \\

\midrule
Ours          & \textbf{71.02} & 70.53 & \textbf{70.97} \\
\bottomrule
\label{table_crack}
\end{tabular}
\end{minipage}
\end{table}

\begin{figure}
	\centering
\includegraphics[width=12 cm]{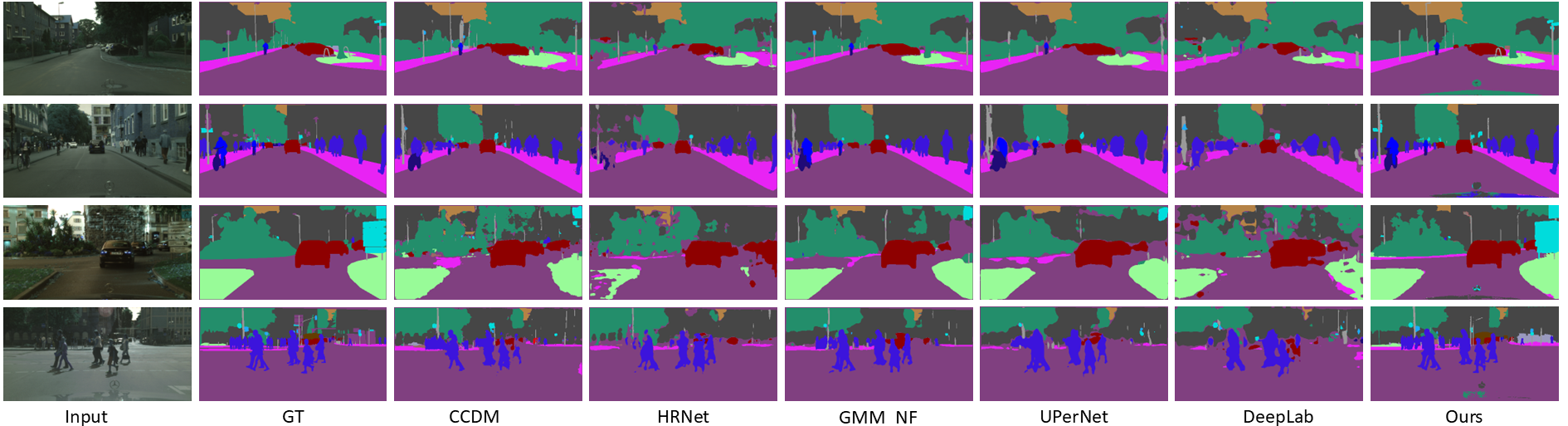}
\caption{Qualitative result from Cityscapes. Following the setting of~\cite{Conghui2025}, all baselines
are trained and tested by 256 × 512 resolution.}
\label{cityscape}
\end{figure}

\subsubsection{ Multi-Class Deterministic Segmentation}
Table~\ref{table_cityscape} summarizes results on Cityscapes. All baseline numbers are from~\cite{Conghui2025}. Although our method is primarily motivated by stochastic segmentation, it also generalizes to deterministic multi-class settings and yields competitive semantic segmentation performance. The model reaches 74.2 IoU, outperforming the strongest baseline in the table while maintaining the same evaluation protocol. For the deterministic task we use MAP inference, and the activated subnet contains 26M parameters. In Fig.~\ref{cityscape}, the predictions exhibit cleaner region consistency and sharper object--background boundaries than the compared outputs.

\subsubsection{Single-Class Curvilinear Deterministic Segmentation}
Table~\ref{table_crack} reports results on Crack500, where the goal is to segment thin crack structures. Baseline results are taken from~\cite{Chen2024}. Our method achieves 71.02 Precision, 70.53 Recall, and 70.97 F1, outperforming prior approaches including CrackFormer~\cite{Liu20212} and JTFN+CIRL~\cite{Chen2024}. These results indicate that the proposed architecture-distribution framework is not restricted to ambiguous segmentation; it also transfers to highly imbalanced curvilinear segmentation and yields competitive deterministic accuracy.

\begin{figure}
	\centering
\includegraphics[width=12 cm]{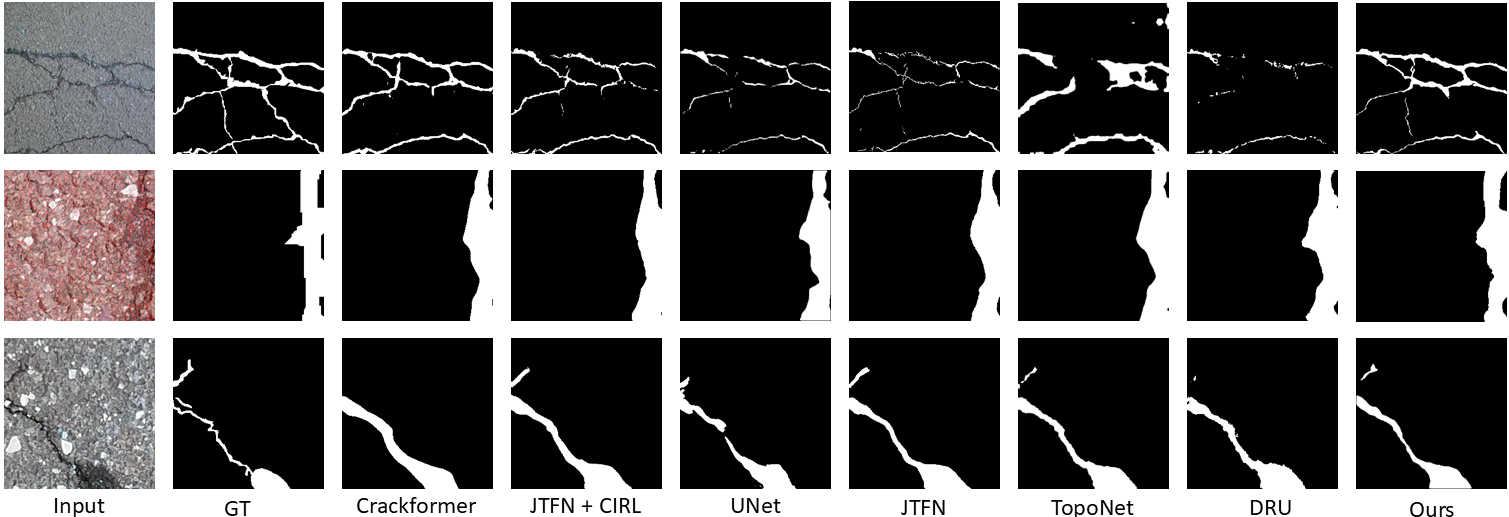}
\caption{Qualitative result from Crack500. Following the setting of~\cite{Chen2024}, all baselines
are trained and tested by 256 × 256 resolution.}
\label{crack}
\end{figure}

\subsection{Inference Efficiency}

\begin{table*}[t]
\caption{Model complexity and inference efficiency. Active Params reports the parameters executed by one sampled forward path (averaged over sampled subnets for Ours). Inference time is measured under a feature-cached protocol for generating $K$ samples from one input at fixed resolution on the same hardware, where deterministic image features are computed once and reused when applicable. Each setting is repeated 10 times and reported as mean and standard deviation.}
\centering
\resizebox{\textwidth}{!}{%
\begin{tabular}{lccccc}
\toprule
\textbf{Method} &
\textbf{Active Params} &
\textbf{Sampling Complexity} &
\textbf{Inference Time$_{K=1}$} &
\textbf{Inference Time$_{K=16}$} &
\textbf{Inference Time$_{K=32}$} \\
\midrule
Prob. U-Net~\cite{Kohl2018}     & 33M  & $O(E + K H)$ & 34.58$_{\pm0.07}$ & 98.4$_{\pm0.6}$  & \textbf{166.3}$_{\pm0.9}$ \\
PHiSeg~\cite{Baumgartner2019}   & 41M  & $O(E + K D)$ & 42.10$_{\pm0.08}$ & 104.7$_{\pm0.8}$ & 172.5$_{\pm1.1}$ \\
cFlow~\cite{Selvan2020}         & 34M  & $O(E + K L_{\mathrm{flow}})$ & 139.84$_{\pm0.11}$ & 635.2$_{\pm1.8}$ & 1163.7$_{\pm2.6}$ \\
GMM\_NF~\cite{Conghui2025}      & 38M  & $O(E + K L_{\mathrm{flow}})$ & 213.02$_{\pm0.14}$ & 1026.4$_{\pm2.4}$ & 1893.8$_{\pm3.5}$ \\
CCDM~\cite{Zbinden2023}         & 30M  & $O(E + KT)$ & 1322.64$_{\pm5.83}$ & 19480.6$_{\pm9.7}$ & 38110.2$_{\pm17.5}$ \\
\midrule
Ours              & 26M(AVG) & $O(E + K D_a)$ & \textbf{28.6}$_{\pm4.2}$ & \textbf{94.8}$_{\pm9.5}$ & 171.9$_{\pm18.7}$ \\
\bottomrule
\end{tabular}%
}
\label{tab:efficiency}
\end{table*}

We compare inference efficiency with representative stochastic segmentation baselines: Prob.\ U-Net~\cite{Kohl2018}, PHiSeg~\cite{Baumgartner2019}, cFlow~\cite{Selvan2020}, GMM\_NF~\cite{Conghui2025}, and CCDM~\cite{Zbinden2023}. Table~\ref{tab:efficiency} reports the active parameter count, the asymptotic sampling complexity when image features are reusable, and the measured feature-cached inference time for generating $K\in\{1,16,32\}$ segmentation hypotheses from a single input.

The cost can be decomposed into a deterministic image-feature term $E$ and a sample-specific stochastic term. For a fixed input, many stochastic segmentation models can cache the encoder features across multiple hypotheses, so increasing $K$ mainly repeats the latent head, decoder, flow, or denoising component rather than the full backbone. The inference times in Table~\ref{tab:efficiency} follow this feature-cached protocol to reflect the standard multi-sample deployment setting. Under the reusable-feature view, Prob.\ U-Net draws one latent code per sample and repeats a low-cost stochastic head, giving $O(E+KH)$; PHiSeg repeats its hierarchical stochastic decoder, giving $O(E+KD)$. Flow-augmented methods (cFlow and GMM\_NF) additionally apply invertible transformations for each sample, resulting in $O(E+K L_{\mathrm{flow}})$, while diffusion-based sampling (CCDM) generates each output through an iterative denoising chain, yielding $O(E+KT)$.

Our method follows the same reusable-feature principle, but the repeated component is an architecture-dependent active path rather than only a final prediction layer. In a cached implementation, the encoder/controller features are computed once, architectures are sampled from $q_{\psi}(a\mid x)$, and each hypothesis executes only its selected path. We denote this per-sample active-path cost by $D_a$, which gives $O(E+K D_a)$. The activated subnet contains 26M parameters on average, while the full supernet contains 387M parameters. Thus, although the model stores a larger candidate bank and incurs additional offline search and training cost, deployment cost is determined by the sampled active path rather than by the complete bank. The $K=1$ inference time is reduced by the complexity-aware evolutionary objective, which favors lower-cost candidate structures. For larger $K$, the inference time remains comparable to CVAE-style baselines while preserving the accuracy and architectural-provenance benefits of architecture sampling. Because different sampled operators instantiate slightly different paths, per-sample inference time varies, which explains the larger standard deviation reported for our method.

\subsection{Ablation Studies}
\label{sec:ablation}

We conduct ablation studies to clarify why both architecture sampling and set-level supervision are necessary for architecture-distribution stochastic segmentation. Unless otherwise stated, ablations are evaluated on LIDC using sampled prediction sets and report HM-IoU, GED, and complementary coverage-oriented metrics.

\subsubsection{Architecture sampler and set-level objective}
We first disentangle the contribution of the architecture sampler from the choice of stochastic objective. Table~\ref{tab:objective_ablation} compares NADS trained with the GED surrogate against alternative set losses and sampling mechanisms at $K=16$. Replacing GED with best-candidate Dice slightly improves its own metric, but substantially degrades HM-IoU and GED, suggesting that best-of-$K$ supervision rewards whether one sample matches an annotation while leaving many sampled architectures weakly constrained. Hungarian matching is stronger, but still below GED because only the matched prediction--annotation pairs receive direct supervision. Conversely, fixed architecture + GED performs poorly, showing that the set-level loss alone cannot create a useful stochastic distribution without a sampling mechanism. A 16-network ensemble is competitive with simple alternatives, but remains below NADS+GED, indicating that weight-shared, input-conditioned architecture sampling is more effective than training independent networks for this setting.

\begin{table}[t]
\centering
\caption{Disentangling the stochastic objective and sampling mechanism on LIDC with $K=16$.}
\resizebox{\textwidth}{!}{%
\begin{tabular}{lccc}
\toprule
\textbf{Variant} & \textbf{HM-IoU} & \textbf{GED} & \textbf{Best-candidate Dice} \\
\midrule
NADS + GED & \textbf{0.658} & \textbf{0.172} & 0.831 \\
NADS + best-candidate Dice & 0.612 & 0.327 & \textbf{0.833} \\
NADS + Hungarian IoU & 0.641 & 0.284 & 0.819 \\
16-network ensemble & 0.623 & 0.276 & 0.817 \\
Fixed architecture + GED & 0.536 & 0.981 & 0.762 \\
\bottomrule
\end{tabular}%
}
\label{tab:objective_ablation}
\end{table}

\subsubsection{Standard losses for stochasticity}
We next replace our set-level GED surrogate with conventional pixel-wise segmentation losses (BCE, Dice), while following a common practice for multi-annotated data: at each iteration we randomly select one mask from the annotation set as the target. We observe a consistent failure mode: the model tends to \emph{lose stochasticity} and the learned architecture distribution collapses to a near-deterministic configuration, which means one operator dominates selection at many searchable positions. As shown in Table~\ref{tab:ablation_loss_evo}, this leads to degraded distributional metrics (HM-IoU$_{16}=0.49$, GED$_{16}=0.89$). This behavior is expected because single-label supervision optimizes a point-estimation objective and provides no explicit incentive to match the multi-modal annotation set or to maintain sample diversity.

\begin{table}[t]
\centering
\caption{Ablations on LIDC ($K{=}16,32$). Standard single-label losses and evolution-only selection both cause the architecture distribution to collapse toward a near-deterministic network, leading to poor distributional metrics.}
\begin{tabular}{lcccc}
\toprule
\textbf{Variant} & \multicolumn{2}{c}{$K{=}16$} & \multicolumn{2}{c}{$K{=}32$} \\
\cmidrule(lr){2-3}\cmidrule(lr){4-5}
 & \textbf{HM-IoU} & \textbf{GED} & \textbf{HM-IoU} & \textbf{GED} \\
\midrule
Dice  & 0.492  & 0.895  & 0.496 & 0.882 \\
BCE   & 0.437  & 0.854  & 0.448 & 0.849 \\
Evolution selection & 0.572 & 1.000 & 0.573 & 0.998 \\
\midrule
Ours  & 0.658 & 0.172 & 0.671 & 0.169 \\
\bottomrule
\end{tabular}
\label{tab:ablation_loss_evo}
\end{table}

\subsubsection{Evolutionary optimization of selection logits}
We also test learning the architecture-selection distribution without gradients: instead of training the controller $\psi$ by backpropagation, we treat the per-position selection logits as an evolvable genotype and update them with an aging evolutionary strategy based on validation fitness. This approach is highly unstable and rapidly collapses to near-deterministic selections, yielding poor distributional performance (Table~\ref{tab:ablation_loss_evo}). We attribute the failure to an optimization mismatch: the fitness is noisy from stochastic sampling and non-stationary due to evolving shared weights $\theta$, so black-box evolution over-exploits early winners and saturates the selection logits, failing to maintain a non-degenerate multi-modal distribution. These results motivate our gradient-based learning of the controller $\psi$ jointly with $\theta$.

\section{Conclusion}
In this work, we introduced architecture distributions as a new stochastic source for segmentation, where diversity is generated by sampling a discrete architecture $a\sim q_\psi(a\mid x)$ and producing each hypothesis through its active path. Unlike latent-code or denoising-noise sources, the support of this stochastic source is a discrete architecture space that can be shaped by evolutionary search before distribution learning, providing a way to better align sampled hypotheses with the empirical annotation distribution while retaining architectural provenance for auditing. We trained the distribution with set-level supervision to avoid averaging collapse and to better match the multi-annotator label set. Empirically, on LIDC-IDRI our model achieves the best performance among compared methods in both distribution matching (GED$_{16}$=0.172) and hypothesis coverage (HM-IoU$_{16}$=0.658). The main limitations are the additional search/training cost required by the candidate bank and the finite support of the architecture space, $|\mathcal A|=\prod_{\ell=1}^{L} C_\ell$. These costs are incurred before deployment, while inference executes only sampled active paths and remains comparable to representative stochastic segmentation baselines. Future work can further explore how to optimize the architecture-distribution support, including search objectives that more directly target distribution matching and calibrated uncertainty.

\input{supplementary}

\bibliographystyle{unsrt}
\bibliography{references}

\end{document}

%% file: supplementary.tex
\clearpage
\appendix
\section*{Supplementary Material}
\addcontentsline{toc}{section}{Supplementary Material}
\section{Implementation detail}

\subsection{Network Architecture}

We insert shape-preserving SearchableBlocks into the decoder feature stream. Concretely, after each decoder upsampling stage, we place $n_{\mathrm{search}}$ searchable blocks, resulting in $L = 4 \times n_{\mathrm{search}}$ searchable positions in total. Each SearchableBlock keeps the input/output tensor shape identical ($\mathbb{R}^{C\times H\times W}\rightarrow \mathbb{R}^{C\times H\times W}$), so it can be dropped into the decoder without changing skip connections or resolution transitions.

Each searchable position $\ell$ maintains a finite candidate set $\mathcal{O}_\ell=\{o_{\ell,1},\dots,o_{\ell,C_\ell}\}$. Candidates are compiled from a compact genotype describing micro-architecture choices, including block family (residual/MBConv/ConvNeXt/identity), kernel size, dilation, expansion ratio, number of layers, optional SE, activation, DropPath, and LayerScale. Unless stated otherwise, we use the same default pool size across positions, and later keep a smaller Top-$N$ subset after evolution.

\subsection{Supernet warmup}
Warmup is used to obtain reliable inherited weights for the large operator bank before evolutionary selection and learning the architecture distribution. Practically, warmup trains only the shared operator weights while disabling any learning of architecture parameters: we freeze all architecture-related parameters and optimize only the supernet weights with AdamW. At each iteration, we create a set of $K_{\mathrm{train}}$ predictions for the same mini-batch by repeating:

\begin{enumerate}
  \item uniformly sample one operator at every searchable position;
  \item execute a single-path forward pass through the instantiated subnetwork.
\end{enumerate}

This produces $K_{\mathrm{train}}$ hypotheses $\{ \hat{y}^{(k)} \}_{k=1}^{K_{\mathrm{train}}}$ stacked as $(B,K_{\mathrm{train}},1,H,W)$, which are optimized with the same set-level stochastic segmentation loss used later in training. Uniform per-position sampling during warmup ensures that all candidates receive gradient updates sufficiently, so the bank becomes broadly functional rather than overfitting to a small subset of operators.

After warmup, we unfreeze architecture parameters and proceed to evolutionary candidate selection and end-to-end training with the reduced Top-$N$ banks.

\subsection{Evolutionary Algorithm}
\label{sec:supp-evo}

\begin{algorithm}[t]
\caption{Aging evolution}
\label{alg:aging-evo}
\begin{algorithmic}[1]
\REQUIRE Warm-started supernet $f_\theta$ with per-position pools $\{\mathcal{O}_\ell\}_{\ell=1}^L$;
search-validation loader $\mathcal{D}_{sv}$; hyper-parameters $(P,S,T,p_{\text{mut}},K_{\text{eval}},B_{\text{eval}},t_{\text{ft}},\eta_{\text{ft}},K_{\text{ft}},\lambda_{\mathrm{ged}},\lambda_{\mathrm{cost}},N,E)$.
\ENSURE Top-$N$ banks $\{\mathcal{B}_\ell\}_{\ell=1}^L$.

\STATE Initialize population $\mathcal{P}\leftarrow \emptyset$, history $\mathcal{H}\leftarrow \emptyset$.
\FOR{$i=1$ to $P$}
  \STATE Sample a random genotype index vector $a^{(i)}$ (biased against identity).
  \STATE Evaluate $\mathrm{fit}(a^{(i)})$ on $\mathcal{D}_{sv}$ with the budgeted protocol.
  \STATE Insert $a^{(i)}$ into $\mathcal{P}$ and $\mathcal{H}$.
\ENDFOR

\FOR{$t=1$ to $T$}
  \STATE Sample a subset $\mathcal{S}\subset\mathcal{P}$ with $|\mathcal{S}|=S$.
  \STATE Parent $a^p \leftarrow \arg\min_{a\in\mathcal{S}} \mathrm{fit}(a)$.
  \STATE Child $a^c \leftarrow \mathrm{Mutate}(a^p, p_{\text{mut}})$ (change 1--3 positions).
  \STATE Evaluate $\mathrm{fit}(a^c)$ on $\mathcal{D}_{sv}$ (lock genotype; optional local fine-tune; metric evaluation; cost proxy).
  \STATE Insert $a^c$ into $\mathcal{P}$ and $\mathcal{H}$.
  \IF{$|\mathcal{P}|>P$}
    \STATE Remove the oldest individual from $\mathcal{P}$ \hfill (aging)
  \ENDIF
\ENDFOR

\STATE Select elites $\mathcal{E}\leftarrow$ the best $E$ individuals from $\mathcal{H}$ by fitness.
\FOR{$\ell=1$ to $L$}
  \STATE Rank candidate indices at position $\ell$ by mean fitness within $\mathcal{E}$.
  \STATE Set $\mathcal{B}_\ell \leftarrow$ Top-$N$ candidates (with deterministic fill if needed).
\ENDFOR
\RETURN $\{\mathcal{B}_\ell\}_{\ell=1}^L$.
\end{algorithmic}
\label{alg:aging}
\end{algorithm}

The evolutionary procedure is used to construct a performance-oriented, position-specific candidate bank for each searchable position. Concretely, we start from a warm-started supernet (Phase~1), run aging evolution on a held-out search-validation split, and finally keep Top-$N$ candidates per position to form the bank used in end-to-end training (Phase~3). The detailed procedure is given in Algorithm~\ref{alg:aging}.

\subsubsection{Search space}
We define $L$ searchable positions in the decoder. At each position $\ell$, a candidate operator is a shape-preserving module ($\mathbb{R}^{C\times H\times W}\rightarrow \mathbb{R}^{C\times H\times W}$), so it can be swapped without changing the global U-Net topology. Each operator is specified by a compact genotype

\begin{equation}
    g = (\texttt{block\_type}, k, d, e, n, r, \texttt{act}, p_{\texttt{dp}}, \gamma_{\texttt{ls}}),
\end{equation}

where block\_type$\in\{$resconv, mbconv, convnext, identity$\}$, $k\in\{3,5,7\}$ is kernel size, $d\in\{1,2\}$ is dilation, $e\in\{2,3,4,6\}$ is expansion ratio, $n\in\{1,2,3\}$ is the number of internal layers (with constraints for some block types), $r\in\{0,0.125,0.25\}$ is SE ratio, act$\in\{$relu, silu, gelu$\}$ is activation, $p_{dp}$ is DropPath rate, and $\gamma_{ls}$ is LayerScale initialization. The full architecture is represented as a length-$L$ index vector $a=[a_1,\ldots,a_L]$, where $a_\ell$ selects one candidate from the local pool at position $\ell$.

\subsubsection{Initialization}
Before evolution, we build an initial pool of $C$ candidate genotypes per position by random sampling  with a fixed RNG seed, while enforcing uniqueness via a hash key of the genotype fields. In our experiments, we set $C=10$ candidates per position. This pool is created once at warmup time and then re-used during evolution and final training.

\subsubsection{Aging evolution}
We use regularized aging evolution with a fixed population size. At each cycle, we sample a small subset tournament, select the best individual as parent, mutate it to produce a child, evaluate the child with a budgeted protocol, then insert the child into the population and remove the oldest individual to keep population size constant. This aging mechanism encourages exploration and reduces premature convergence.

Given a parent index vector $a$, we mutate it with rate $p_{\text{mut}}$ by randomly changing 1 to 3 positions; for each chosen position, the current index is replaced with another valid index from that position’s pool excluding the current one. This mutation has low computational cost and keeps the global topology unchanged.

\subsubsection{Fitness evaluation}
To evaluate an individual $a$, we lock the supernet to the genotype specified by $a$, instantiate the corresponding single-path subnet, and compute its HM-IoU, GED, and complexity-aware fitness on the search-validation loader under a fixed evaluation budget: for each input, we sample architectures from the induced distribution over the restricted candidate sets and execute a single-path forward pass to obtain a prediction set. We generate $K_{\text{eval}}$ sampled predictions per input to interface with the same set-based evaluation code, and we limit evaluation to at most $B_{\text{eval}}$ images to control cost.

Furthermore, we perform a short local fine-tune on the search-validation stream to improve genotype distinguishability. We fine-tune only the shared weights for $t_{\text{ft}}$ steps with the locked genotype, then restore the original shared weights from a snapshot to avoid contaminating the supernet. Architecture parameters remain frozen throughout fine-tuning. During this local fine-tune, we keep the genotype locked in the same way and sample architectures as above, while updating only the shared weights. We combine distributional metrics with a complexity proxy:

\begin{equation}
\mathrm{fit}(a) \;=\; (1-\mathrm{HM\mbox{-}IoU}) \;+\; \lambda_{\mathrm{ged}}\mathrm{GED} \;+\; \lambda_{\mathrm{cost}}\mathrm{Cost}(a),
\end{equation}
where $\mathrm{Cost}(a)$ accumulates per-position genotype complexity scores:

\begin{equation}
\mathrm{Cost}(a) \;=\; \sum_{\ell=1}^{L} \mathrm{cost}(g_{\ell,a_\ell}), \quad
\mathrm{cost}(g) \propto (k^2)\,(C^2)\,\max(n,1)\cdot \eta(g),
\end{equation}
and $\eta(g)$ applies small multiplicative factors for MBConv/ConvNeXt expansion, dilation, and SE usage . Lower fitness is better.

\subsubsection{\texorpdfstring{Top-$N$ extraction}{Top-N extraction}.}
After evolution, we take the best $E$ elites from the entire history ranked by fitness. For each position, we compute the mean fitness of each candidate index among elites, rank candidates by this mean, and keep the Top-$N$ genotypes. If fewer than $N$ unique indices appear among elites at a position, we fill the bank with unused pool candidates and with additional randomly sampled non-identity genotypes. The resulting Top-$N$ per-position banks are then used to instantiate the final training model.

\subsection{Parameter Settings}
This section summarizes the default hyper-parameters used in our experiments, following the released implementation.

\subsubsection{Dataset setting}
\paragraph{LIDC-IDRI}
is a thoracic CT dataset for lung cancer screening/diagnosis, where lesions are
annotated by multiple radiologists, making it suitable for studying annotation ambiguity. We partition the dataset into three disjoint subsets: train for supernet warmup and final training, search-val used only for evolutionary fitness evaluation and training-time monitoring, and test sealed until the final report. Concretely, we set search\_val\_split=0.10 and test\_split=0.20, and only apply paired geometric augmentation on the training subset (random flips and $90^\circ$ rotations). 

\paragraph{Crack500}
is a pixel-wise pavement crack segmentation dataset. Following the standard protocol, the
dataset is provided as 500 high-resolution images and further processed into 3,368 cropped samples
of size $360\times 640$. We use the official split with 1,897 training images, 347 validation images,
and 1,124 testing images. During training, we apply horizontal flipping, random cropping, and random
rotations of $90^\circ/180^\circ/270^\circ$; additionally, all training samples are cropped to
$256\times 256$ for optimization and all methods are trained/tested at $256\times 256$ resolution.

\paragraph{Cityscapes}
provides fine-grained pixel annotations for urban driving scenes with 19 semantic
classes, serving as a representative multi-class deterministic segmentation benchmark. We use the
official split containing 2,975 training images and 500 validation images at native resolution
$512\times 1024$. Following the referenced setting, we use a resized resolution of $256\times 512$ for
both training and evaluation.

\subsubsection{Phase 1: Warmup}
We warm up the shared supernet weights using SPOS-style single-path uniform sampling: at each step, each searchable position randomly selects one operator, and we aggregate $K$ such stochastic forward passes to form a prediction set for set-level supervision. During warmup, architecture parameters $\psi$ are frozen and only inherited weights $\theta$ are updated with AdamW. We use warmup\_steps=500, warmup\_K=4, and warmup\_lr=$2\times10^{-4}$.

\subsubsection{Phase 2: Aging Evolution}
Unless otherwise stated, we use a pool size per position of $C=10$ with a retained bank size per position of $N=5$; a population size of $P=40$, tournament size of $S=10$, $T=120$ cycles, and mutation probability $p_{\text{mut}}=0.30$; an elite keep of $E=30$ for Top-$N$ extraction; evaluation settings of $K_{\text{eval}}=16$ forward passes per input and $B_{\text{eval}}=120$ images; local fine-tuning for $t_{\text{ft}}=30$ steps with learning rate $5\times 10^{-5}$ and $K_{\text{ft}}=4$ passes per step; and fitness weights $\lambda_{\mathrm{ged}}=1.0$ and $\lambda_{\mathrm{cost}}=0.02$.

\subsubsection{Phase 3: Training}
We train the final NADS model for epochs=200 with batch\_size=8 and num\_workers=4. We use two AdamW optimizers: one for inherited weights $\theta$ (lr\_theta=$1\times10^{-4}$, weight\_decay=$1\times10^{-4}$) and one for architecture parameters $\psi$ (lr\_psi=$3\times10^{-4}$, weight\_decay=0). We freeze $\psi$ for the first warmup\_epochs=$0.1\times \texttt{epochs}$ and then jointly optimize $(\theta,\psi)$.
For the training objective, we use a diversity weight $\beta=1.0$ in $\mathcal{L}_{\mathrm{set}}$, a small auxiliary-loss weight $\lambda_{\mathrm{aux}}=0.1$, and an architecture-regularizer weight $\lambda_{\mathrm{arch}}=0.01$ that is annealed toward $0$ together with the temperature $\tau$.

We employ a straight-through Gumbel-Softmax relaxation with temperature $\tau$ during training and progressively anneal it to encourage more discrete behavior. Concretely, $\psi$ is frozen for the first $0.1E$ epochs (with $E$ the total number of epochs), after which $\tau$ is linearly annealed from $5.0$ to $0.5$:
\[
\tau(e)=
\begin{cases}
5.0, & e < 0.1E,\\[2pt]
\max\!\left(0.5,\; 5.0 - 4.5\,\dfrac{e-0.1E}{0.9E}\right), & e \ge 0.1E,
\end{cases}
\]
where $e$ is the current epoch. The entropy regularizer coefficient and the uniform-mix probability are also decayed over training. For stochastic prediction, we sample K\_pred=16 architectures per input (threshold at 0.5, consistent with Eq.~(6)) and evaluate with tau\_eval=0.7; we cap GED/HM-IoU evaluation to max\_images\_for\_ged=200 images when monitoring.

\section{Visualization of operator probabilities}

\begin{figure}
	\centering
\includegraphics[width=12 cm]{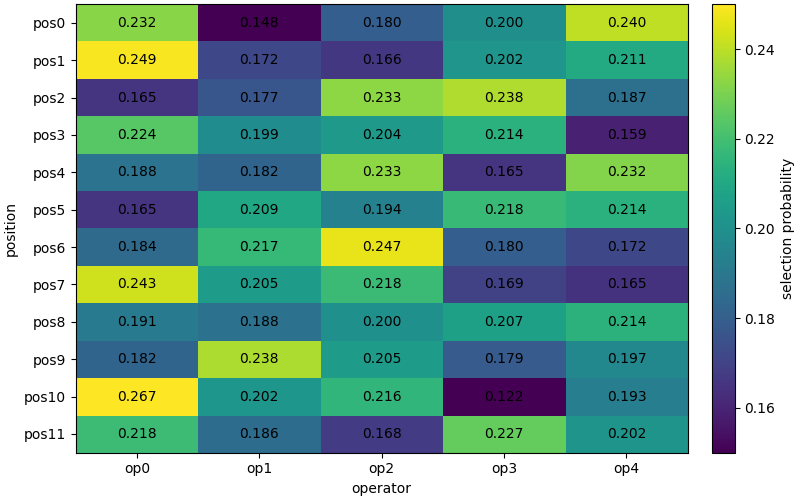}
\caption{Per-position operator selection probabilities for a random input.}
\label{heatmap}
\end{figure}

To verify that the learned architecture controller produces a non-degenerate input-conditioned distribution rather than collapsing to a single deterministic path, we visualize the operator selection probabilities at every searchable position for a randomly sampled input in Fig.~\ref{heatmap}. The heatmap indicates that different positions exhibit different preferences over operators, suggesting depth- and stage-dependent operator roles. No single operator dominates all positions, which supports the use of conditional computation and stochastic sampling at inference. This qualitative evidence complements the quantitative gains: the controller learns a structured, position-aware mixture over candidate operators, enabling diverse hypotheses while still executing only one sampled path per forward pass.

\section{Training time}

\begin{table}[t]
\centering
\caption{Training time per epoch for representative stochastic segmentation baselines and the proposed method. NADS requires the largest training time because it optimizes a full supernet with a large candidate operator bank.}
\label{tab:train_time_a100}
\begin{tabular}{l c}
\toprule
\textbf{Method} & \textbf{Time/epoch (min:sec)} \\
\midrule
Prob.\ U-Net \cite{Kohl2018} & 2:28 \\
PHiSeg \cite{Baumgartner2019} & 3:13 \\
GMM\_NF \cite{Conghui2025} & 5:52 \\
CCDM \cite{Zbinden2023} & 2:35 \\
\midrule
Ours & 9:42 \\
\bottomrule
\end{tabular}
\end{table}

\paragraph{Justification.}
Table~\ref{tab:train_time_a100} compares the training time of NADS with representative stochastic segmentation baselines. NADS has higher training cost because it optimizes a supernet containing a large candidate operator bank. Although inference executes only a single sampled path, training must repeatedly sample architectures and update the inherited weights of many candidate operators so that the search space is sufficiently trained for reliable selection and distribution learning. Consequently, each epoch takes longer (9m42s) than the compared stochastic segmentation baselines. This additional cost is incurred during offline training, while deployment benefits from the resulting distributional accuracy and calibrated predictive diversity.

\section{Capability evaluation}

We ablate the number of searchable positions ($L$) and the per-position retained operator budget (Top-$N$) to understand the capacity–performance trade-off under small epoch at Table~\ref{tab:positions_topk}. In this experiment, we applied 100 epochs per setting. Each setting are trained and evaluated 3 times to take average. As shown in Table~\ref{tab:positions_topk}, increasing $L$ from 4 to 12 generally improves both accuracy and distributional fidelity under a moderate $K$ (with Top-5, HM-IoU increases from 0.613 to 0.622, while GED decreases from 0.241 to 0.231), indicating that more positions provide additional stage-wise flexibility for input-conditioned operator selection and better alignment with multi-annotation uncertainty. In contrast, using an overly large retained set (Top-10) can be suboptimal, likely because low-ranked operators add optimization noise and weaken the quality of the learned operator distribution, leading to degraded HM-IoU/GED in some settings.

The method also remains effective under small capacity. With $L{=}4$ and Top-3 per position, corresponding to 12 available operator choices in total, the model achieves HM-IoU 0.606 and GED 0.261. This suggests that a large operator pool is not required for competitive distributional quality. The observation is consistent with the intuition that, when the retained per-position budget $N$ is at least the number of dominant annotation modes, the model can allocate enough distinct operators to cover the major modes; we treat this as an empirical observation rather than a guarantee. Overall, $L{=}12$ with Top-5 yields the best trade-off among the evaluated configurations and is used as the default configuration.

\begin{table*}[t]
\centering
\caption{Ablation on the number of searchable positions and the retained candidate bank size (Top-$N$ per position).}
\label{tab:positions_topk}
\begin{tabular*}{\textwidth}{@{\extracolsep{\fill}} c cc cc cc}
\toprule
\multirow{2}{*}{\#Positions} &
\multicolumn{2}{c}{Top-$3$} &
\multicolumn{2}{c}{Top-$5$} &
\multicolumn{2}{c}{Top-$10$} \\
\cmidrule(lr){2-3}\cmidrule(lr){4-5}\cmidrule(lr){6-7}
& HM-IoU & GED
& HM-IoU & GED
& HM-IoU & GED \\
\midrule
4  & 0.606 & 0.261 & 0.613 & 0.241 & 0.608 & 0.245 \\
8  & 0.609 & 0.241 & 0.617 & 0.237 & 0.600 & 0.264 \\
12 & 0.611 & 0.241 & \textbf{0.622} & \textbf{0.231} & 0.608 & 0.255 \\
\bottomrule
\end{tabular*}
\end{table*}

\section{\texorpdfstring{Temperature $\tau$ evaluation}{Temperature tau evaluation}}

Our architecture controller parameterizes an input-conditioned categorical distribution $q_{\psi}(a\mid x)=\prod_{\ell=1}^{L} q_{\psi}(a_\ell\mid x)$ over discrete operator choices, from which we sample architectures to generate $K$ hypotheses at inference time. Since directly optimizing discrete architectural choices is non-differentiable, during training we adopt the straight-through Gumbel-Softmax relaxation with temperature $\tau$ (Eq.~(7) in main paper) to obtain a relaxed one-hot vector $\tilde a_\ell$. As stated in our method, we progressively reduce $\tau$ in later stages to encourage sharper distributions and more discrete behavior, better matching the discrete sampling used at inference. 

We compare three strategies during training: Fixed-High temperature, Fixed-Low temperature, and Annealed temperature (start high and progressively decrease to a low value). We report distributional quality metrics together with an optimization diagnostic, the average architecture entropy $H(q_{\psi}(a_\ell\mid x))$. All experiments are trained for 100 epochs.

\begin{table*}[t]
\centering
\caption{Reference numbers for ablating the Gumbel-Softmax temperature $\tau$ and annealing strategy under straight-through sampling. We report the average gate entropy $\mathcal{H}$ across all searchable positions. $\mathcal{H}_{0.1T}$ and $\mathcal{H}_{T}$ denote the average gate entropy measured at 10\% and 100\% of training, respectively.}
\label{tab:tau_ablation_ref}
\begin{tabular*}{\textwidth}{@{\extracolsep{\fill}} l c cc cc}
\toprule
Setting & $\tau$ schedule
& $\mathcal{H}_{0.1T}$  & $\mathcal{H}_{T}$ 
& HM-IoU  & GED  \\
\midrule
Fixed high $\tau$  & $\tau=5.0$
& 1.46 & 1.38
& 0.615 & 0.242
 \\
Fixed low $\tau$   & $\tau=0.5$
& 0.62 & 0.25
& 0.607 & 0.252
 \\
Piecewise anneal & $5.0 \rightarrow 0.5$
& 1.44 & 0.53
& \textbf{0.630} & \textbf{0.233}
 \\
\bottomrule
\end{tabular*}

\end{table*}

All results are summarized in Table~\ref{tab:tau_ablation_ref}. Fixed-High keeps the relaxation overly soft, so the controller learns under a ``mixture-like'' proxy while inference relies on discrete architecture sampling. This train--test mismatch typically hurts both segmentation quality and set-level matching. Fixed-Low makes the relaxation near-discrete from the beginning; the straight-through estimator becomes high-variance and the controller tends to prematurely concentrate on a few operators, leading to reduced entropy, even with the negative-entropy regularizer $\mathcal{R}(q_\psi)$. In contrast, the annealed schedule provides a stable exploration--exploitation trade-off: early training benefits from smoother gradients and exploration (higher entropy), while later training becomes more discrete to align with inference-time sampling, yielding the best distributional metrics.